# Deep Variational Canonical Correlation Analysis

**Weiran Wang** [1]  **Xinchen Yan** [2]  **Honglak Lee** [2]  **Karen Livescu** [1]

## Abstract

We present deep variational canonical correlation analysis (VCCA), a deep multi-view learning model that extends the latent variable model interpretation of linear CCA to nonlinear observation models parameterized by deep neural networks. We derive variational lower bounds of the data likelihood by parameterizing the posterior probability of the latent variables from the view that is available at test time. We also propose a variant of VCCA called VCCA-private that can, in addition to the "common variables" underlying both views, extract the "private variables" within each view, and disentangles the shared and private information for multi-view data without hard supervision. Experimental results on real-world datasets show that our methods are competitive across domains.

## 1. Introduction

In the multi-view representation learning setting, we have multiple views (types of measurements) of the same underlying signal, and the goal is to learn useful features of each view using complementary information contained in both views. The learned features should uncover the common sources of variation in the views, which can be helpful for exploratory analysis or for downstream tasks.

A classical approach is canonical correlation analysis (CCA, Hotelling, 1936) and its nonlinear extensions, including the kernel extension (Lai & Fyfe, 2000; Akaho, 2001; Melzer et al., 2001; Bach & Jordan, 2002) and the deep neural network (DNN) extension (Andrew et al., 2013; Wang et al., 2015b). CCA projects two random vectors $\mathbf{x} \in \mathbb{R}^{d_x}$ and $\mathbf{y} \in \mathbb{R}^{d_y}$ into a lower-dimensional subspace so that the projections are maximally correlated. There is a probabilistic latent variable model interpretation of linear CCA as shown in Figure 1 (left). Assuming that $\mathbf{x}$ and $\mathbf{y}$ are linear functions of some random variable $\mathbf{z} \in \mathbb{R}^{d_z}$ where $d_z \leq \min(d_x, d_y)$, and that the prior distribution $p(\mathbf{z})$ and conditional distributions $p(\mathbf{x}|\mathbf{z})$ and $p(\mathbf{y}|\mathbf{z})$ are Gaussian, Bach & Jordan (2005) showed that $\mathbb{E}[\mathbf{z}|\mathbf{x}]$ (resp. $\mathbb{E}[\mathbf{z}|\mathbf{y}]$) lives in the same space as the linear CCA projection for $\mathbf{x}$ (resp. $\mathbf{y}$).

[1]Toyota Technological Institute at Chicago, Chicago, IL 60637, USA [2]University of Michigan, Ann Arbor, MI 48109, USA. Correspondence to: Weiran Wang <weiranwang@ttic.edu>.

This generative interpretation of CCA is often lost in its nonlinear extensions. For example, in deep CCA (DCCA, Andrew et al., 2013), one extracts nonlinear features from the original inputs of each view using two DNNs, $\mathbf{f}$ for $\mathbf{x}$ and $\mathbf{g}$ for $\mathbf{y}$, so that the canonical correlation of the DNN outputs (measured by a linear CCA with projection matrices $\mathbf{U}$ and $\mathbf{V}$) is maximized. Formally, given a dataset of $N$ pairs of observations $(\mathbf{x}_1, \mathbf{y}_1), \ldots, (\mathbf{x}_N, \mathbf{y}_N)$ of the random vectors $(\mathbf{x}, \mathbf{y})$, DCCA optimizes

$$\max_{\mathbf{W_f}, \mathbf{W_g}, \mathbf{U}, \mathbf{V}} \quad \operatorname{tr}\left(\mathbf{U}^\top \mathbf{f}(\mathbf{X}) \mathbf{g}(\mathbf{Y})^\top \mathbf{V}\right) \quad (1)$$

$$\text{s.t.} \ \mathbf{U}^\top \left(\mathbf{f}(\mathbf{X})\mathbf{f}(\mathbf{X})^\top\right) \mathbf{U} = \mathbf{V}^\top \left(\mathbf{g}(\mathbf{Y})\mathbf{g}(\mathbf{Y})^\top\right) \mathbf{V} = N\mathbf{I},$$

where $\mathbf{W_f}$ (resp. $\mathbf{W_g}$) denotes the weight parameters of $\mathbf{f}$ (resp. $\mathbf{g}$), and $\mathbf{f}(\mathbf{X}) = [\mathbf{f}(\mathbf{x}_1), \ldots, \mathbf{f}(\mathbf{x}_N)]$, $\mathbf{g}(\mathbf{Y}) = [\mathbf{g}(\mathbf{y}_1), \ldots, \mathbf{g}(\mathbf{y}_N)]$.

DCCA has achieved good performance across several domains (Wang et al., 2015b;a; Lu et al., 2015; Yan & Mikolajczyk, 2015). However, a disadvantage of DCCA is that it does not provide a model for generating samples from the latent space. Although Wang et al. (2015b)'s deep canonically correlated autoencoders (DCCAE) variant optimizes the combination of an autoencoder objective (reconstruction errors) and the canonical correlation objective, the authors found that in practice, the canonical correlation term often dominate the reconstruction terms in the objective, and therefore the inputs are not reconstructed well. At the same time, optimization of the DCCA/DCCAE objectives is challenging due to the constraints that couple all training samples.

The main contribution of this paper is a new deep multi-view learning model, deep variational CCA (VCCA), which extends the latent variable interpretation of linear CCA to nonlinear observation models parameterized by DNNs. Computation of the marginal data likelihood and inference of the latent variables are both intractable under this model. Inspired by variational autoencoders

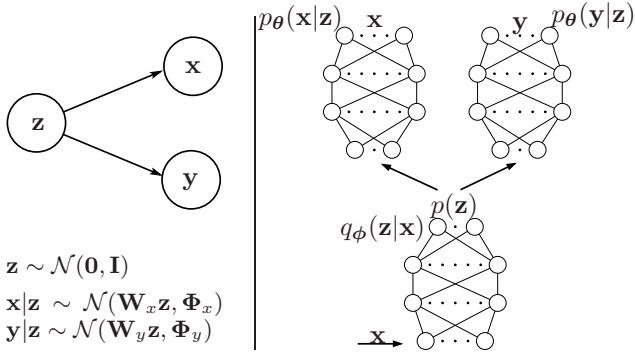

$\mathbf{z} \sim \mathcal{N}(\mathbf{0}, \mathbf{I})$

$\mathbf{x} | \mathbf{z} \sim \mathcal{N}(\mathbf{W}_x \mathbf{z}, \boldsymbol{\Phi}_x)$

$\mathbf{y} | \mathbf{z} \sim \mathcal{N}(\mathbf{W}_y \mathbf{z}, \boldsymbol{\Phi}_y)$

*Figure 1.* Left: Probabilistic latent variable interpretation of CCA (Bach & Jordan, 2005). Right: Deep variational CCA.

(VAE, Kingma & Welling, 2014), we parameterize the posterior distribution of the latent variables given an input view, and derive variational lower bounds of the data likelihood, which is further approximated by Monte Carlo sampling. With the reparameterization trick, sampling for the Monte Carlo approximation is trivial and all DNN weights in VCCA can be optimized jointly via stochastic gradient descent, using unbiased gradient estimates from small minibatches. Interestingly, VCCA is related to multi-view autoencoders (Ngiam et al., 2011), with additional regularization on the posterior distribution.

We also propose a variant of VCCA called VCCA-private that can, in addition to the "common variables" underlying both views, extract the "private variables" within each view. We demonstrate that VCCA-private is able to disentangle the shared and private information for multi-view data without hard supervision. Last but not least, as generative models, VCCA and VCCA-private enable us to obtain high-quality samples for the input of each view.

## 2. Variational CCA

The probabilistic latent variable model of CCA (Bach & Jordan, 2005) defines the following joint distribution over the random variables $(\mathbf{x}, \mathbf{y})$:

$$p(\mathbf{x}, \mathbf{y}, \mathbf{z}) = p(\mathbf{z}) p(\mathbf{x}|\mathbf{z}) p(\mathbf{y}|\mathbf{z}), \qquad (2)$$

$$p(\mathbf{x}, \mathbf{y}) = \int p(\mathbf{x}, \mathbf{y}, \mathbf{z}) d\mathbf{z}.$$

The assumption here is that, conditioned on the latent variables $\mathbf{z} \in \mathbb{R}^{d_z}$, the two views $\mathbf{x}$ and $\mathbf{y}$ are independent. Classical CCA is obtained by assuming that the observation models $p(\mathbf{x}|\mathbf{z})$ and $p(\mathbf{y}|\mathbf{z})$ are linear, as shown in Figure 1 (left). However, linear observation models have limited representation power. In this paper, we consider nonlinear observation models $p_{\boldsymbol{\theta}}(\mathbf{x}|\mathbf{z}; \boldsymbol{\theta}_x)$ and $p_{\boldsymbol{\theta}}(\mathbf{y}|\mathbf{z}; \boldsymbol{\theta}_y)$, parameterized by $\boldsymbol{\theta}_x$ and $\boldsymbol{\theta}_y$ respectively, which can be the collections of weights of DNNs. In this case, the marginal likelihood $p_{\boldsymbol{\theta}}(\mathbf{x}, \mathbf{y})$ does not have a closed form, and the inference problem $p_{\boldsymbol{\theta}}(\mathbf{z}|\mathbf{x})$—the problem of inferring the latent variables given one of the views—is also intractable.

Inspired by Kingma & Welling (2014)'s work on variational autoencoders (VAE), we approximate $p_{\boldsymbol{\theta}}(\mathbf{z}|\mathbf{x})$ with the conditional density $q_{\boldsymbol{\phi}}(\mathbf{z}|\mathbf{x}; \boldsymbol{\phi}_z)$, where $\boldsymbol{\phi}_z$ is the collection of parameters of another DNN.[1] We can derive a lower bound on the marginal data log-likelihood using $q_{\boldsymbol{\phi}}(\mathbf{z}|\mathbf{x})$: (see the full derivation in Appendix A)

$$\log p_{\boldsymbol{\theta}}(\mathbf{x}, \mathbf{y}) \geq \mathcal{L}(\mathbf{x}, \mathbf{y}; \boldsymbol{\theta}, \boldsymbol{\phi}) := -D_{KL}(q_{\boldsymbol{\phi}}(\mathbf{z}|\mathbf{x})||p(\mathbf{z}))$$
$$+ \mathbb{E}_{q_{\boldsymbol{\phi}}(\mathbf{z}|\mathbf{x})} \left[ \log p_{\boldsymbol{\theta}}(\mathbf{x}|\mathbf{z}) + \log p_{\boldsymbol{\theta}}(\mathbf{y}|\mathbf{z}) \right] \qquad (3)$$

where $D_{KL}(q_{\boldsymbol{\phi}}(\mathbf{z}|\mathbf{x})||p(\mathbf{z}))$ denotes the KL divergence between the approximate posterior $q_{\boldsymbol{\phi}}(\mathbf{z}|\mathbf{x})$ and the prior $q(\mathbf{z})$ for the latent variables. VCCA maximizes this variational lower bound on the data log-likelihood on the training set:

$$\max_{\boldsymbol{\theta}, \boldsymbol{\phi}} \frac{1}{N} \sum_{i=1}^{N} \mathcal{L}(\mathbf{x}_i, \mathbf{y}_i; \boldsymbol{\theta}, \boldsymbol{\phi}). \qquad (4)$$

**The KL divergence term** When the parameterization $q_{\boldsymbol{\phi}}(\mathbf{z}|\mathbf{x})$ is chosen properly, this term can be computed exactly in closed form. Let the variational approximate posterior be a multivariate Gaussian with diagonal covariance. That is, for a sample pair $(\mathbf{x}_i, \mathbf{y}_i)$, we have

$$\log q_{\boldsymbol{\phi}}(\mathbf{z}_i|\mathbf{x}_i) = \log \mathcal{N}(\mathbf{z}_i; \boldsymbol{\mu}_i, \boldsymbol{\Sigma}_i),$$
$$\boldsymbol{\Sigma}_i = \text{diag}\left(\sigma_{i1}^2, \ldots, \sigma_{id_z}^2\right),$$

where the mean $\boldsymbol{\mu}_i$ and covariance $\boldsymbol{\Sigma}_i$ are outputs of an encoding DNN $\mathbf{f}$ (and thus $[\boldsymbol{\mu}_i, \boldsymbol{\Sigma}_i] = \mathbf{f}(\mathbf{x}_i; \boldsymbol{\phi}_z)$ are *deterministic* nonlinear functions of $\mathbf{x}_i$). In this case, we have

$$D_{KL}(q_{\boldsymbol{\phi}}(\mathbf{z}_i|\mathbf{x}_i)||p(\mathbf{z}_i)) = -\frac{1}{2} \sum_{j=1}^{d_z} \left( 1 + \log \sigma_{ij}^2 - \sigma_{ij}^2 - \mu_{ij}^2 \right).$$

**The expected log-likelihood term** The second term of (3) corresponds to the expected data log-likelihood under the approximate posterior distribution. Though still intractable, this term can be approximated by Monte Carlo sampling: We draw $L$ samples $\mathbf{z}_i^{(l)} \sim q_{\boldsymbol{\phi}}(\mathbf{z}_i|\mathbf{x}_i)$ where

$$\mathbf{z}_i^{(l)} = \boldsymbol{\mu}_i + \boldsymbol{\Sigma}_i \boldsymbol{\epsilon}^{(l)}, \quad \text{where } \boldsymbol{\epsilon}^{(l)} \sim \mathcal{N}(\mathbf{0}, \mathbf{I}), \ l = 1, \ldots, L,$$

and have

$$\mathbb{E}_{q_{\boldsymbol{\phi}}(\mathbf{z}_i|\mathbf{x}_i)} \left[ \log p_{\boldsymbol{\theta}}(\mathbf{x}_i|\mathbf{z}_i) + \log p_{\boldsymbol{\theta}}(\mathbf{y}_i|\mathbf{z}_i) \right] \approx$$
$$\frac{1}{L} \sum_{l=1}^{L} \log p_{\boldsymbol{\theta}} \left( \mathbf{x}_i|\mathbf{z}_i^{(l)} \right) + \log p_{\boldsymbol{\theta}} \left( \mathbf{y}_i|\mathbf{z}_i^{(l)} \right). \qquad (5)$$

We provide a sketch of VCCA in Figure 1 (right).

---

[1] For notational simplicity, we denote by $\boldsymbol{\theta}$ the parameters associated with the model probabilities $p_{\boldsymbol{\theta}}(\cdot)$, and $\boldsymbol{\phi}$ the parameters associated with the variational approximate probabilities $q_{\boldsymbol{\phi}}(\cdot)$, and often omit specific parameters inside the probabilities.

**Connection to multi-view autoencoder (MVAE)** If we use the Gaussian observation models

$$\log p_{\boldsymbol{\theta}}(\mathbf{x}|\mathbf{z}) = \log \mathcal{N}(\mathbf{g}_x(\mathbf{z}; \boldsymbol{\theta}_x), \mathbf{I}),$$
$$\log p_{\boldsymbol{\theta}}(\mathbf{y}|\mathbf{z}) = \log \mathcal{N}(\mathbf{g}_y(\mathbf{z}; \boldsymbol{\theta}_y), \mathbf{I}),$$

we observe that $\log p_{\boldsymbol{\theta}}\left(\mathbf{x}_i|\mathbf{z}_i^{(l)}\right)$ and $\log p_{\boldsymbol{\theta}}\left(\mathbf{y}_i|\mathbf{z}_i^{(l)}\right)$ measure the $\ell_2$ reconstruction errors of each view's inputs from samples $\mathbf{z}_i^{(l)}$ using the two DNNs $\mathbf{g}_x$ and $\mathbf{g}_y$ respectively. In this case, maximizing $\mathcal{L}(\mathbf{x}, \mathbf{y}; \boldsymbol{\theta}, \boldsymbol{\phi})$ is equivalent to

$$\min_{\boldsymbol{\theta}, \boldsymbol{\phi}} \frac{1}{N} \sum_{i=1}^{N} D_{KL}(q_{\boldsymbol{\phi}}(\mathbf{z}_i|\mathbf{x}_i)||p(\mathbf{z}_i)) \qquad (6)$$

$$+ \frac{1}{2NL} \sum_{i,l} \left\| \mathbf{x}_i - \mathbf{g}_x(\mathbf{z}_i^{(l)}; \boldsymbol{\theta}_x) \right\|^2 + \left\| \mathbf{y}_i - \mathbf{g}_y(\mathbf{z}_i^{(l)}; \boldsymbol{\theta}_y) \right\|^2$$

s.t. $\mathbf{z}_i^{(l)} = \boldsymbol{\mu}_i + \boldsymbol{\Sigma}_i \boldsymbol{\epsilon}^{(l)}$, where $\boldsymbol{\epsilon}^{(l)} \sim \mathcal{N}(\mathbf{0}, \mathbf{I})$, $l = 1, \ldots, L$.

Now, consider the case of $\boldsymbol{\Sigma}_i \to \mathbf{0}$, and we have $\mathbf{z}_i^{(l)} \to \boldsymbol{\mu}_i$ which is a deterministic function of $\mathbf{x}$ (and there is no need for sampling). In the limit, the second term of (6) becomes

$$\frac{1}{2N} \sum_{i=1}^{N} \|\mathbf{x}_i - \mathbf{g}_x(\mathbf{f}(\mathbf{x}_i))\|^2 + \|\mathbf{y}_i - \mathbf{g}_y(\mathbf{f}(\mathbf{x}_i))\|^2, \qquad (7)$$

which is the objective of the multi-view autoencoder (MVAE, Ngiam et al., 2011). Note, however, that $\boldsymbol{\Sigma}_i \to \mathbf{0}$ is prevented by the VCCA objective as it results in a large penalty in $D_{KL}(q_{\boldsymbol{\phi}}(\mathbf{z}_i|\mathbf{x}_i)||p(\mathbf{z}_i))$. Compared with the MVAE objective, in the VCCA objective we are creating $L$ different "noisy" versions of the latent representation and enforce that these versions reconstruct the original inputs well. The "noise" distribution (the variances $\boldsymbol{\Sigma}_i$) are also learned and regularized by the KL divergence $D_{KL}(q_{\boldsymbol{\phi}}(\mathbf{z}_i|\mathbf{x}_i)||p(\mathbf{z}_i))$. Using the VCCA objective, we expect to learn different representations from those of MVAE, due to these regularization effects.

## 2.1. Extracting private variables

A potential disadvantage of VCCA is that it assumes the common latent variables $\mathbf{z}$ are sufficient to generate the views, which can be too restrictive in practice. Consider the example of audio and articulatory measurements as two views for speech. Although the transcription is a common variable behind the views, it combines with the physical environment and the vocal tract anatomy to generate the individual views. In other words, there might be large variations in the input space that can not be explained by the common variables, making the objective (3) hard to optimize. It may then be beneficial to explicitly model the private variables within each view. See "The effect of private variables on reconstructions" in Section 4.1 for an illustration of this intuition.

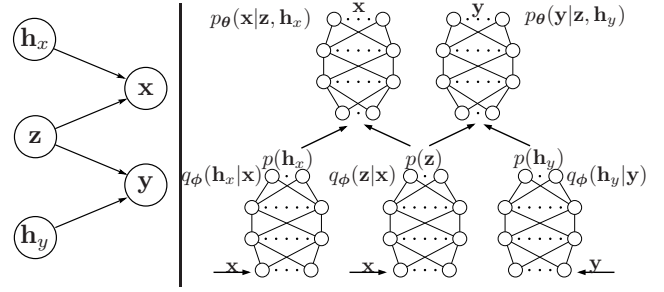

*Figure 2.* VCCA-private: variational CCA with view-specific private variables.

We propose a second model, whose graphical model is shown in Figure 2, that we refer to as VCCA-private. We introduce two sets of hidden variables $\mathbf{h}_x \in \mathbb{R}^{d_{h_x}}$ and $\mathbf{h}_y \in \mathbb{R}^{d_{h_y}}$ to explain the aspects of $\mathbf{x}$ and $\mathbf{y}$ not captured by the common variables $\mathbf{z}$. Under this model, the data likelihood is defined by

$$p_{\boldsymbol{\theta}}(\mathbf{x}, \mathbf{y}, \mathbf{z}, \mathbf{h}_x, \mathbf{h}_y) =$$
$$p(\mathbf{z})p(\mathbf{h}_x)p(\mathbf{h}_y)p_{\boldsymbol{\theta}}(\mathbf{x}|\mathbf{z}, \mathbf{h}_x; \boldsymbol{\theta}_x)p_{\boldsymbol{\theta}}(\mathbf{y}|\mathbf{z}, \mathbf{h}_y; \boldsymbol{\theta}_y),$$

$$p_{\boldsymbol{\theta}}(\mathbf{x}, \mathbf{y}) = \iiint p_{\boldsymbol{\theta}}(\mathbf{x}, \mathbf{y}, \mathbf{z}, \mathbf{h}_x, \mathbf{h}_y) d\mathbf{z} \, d\mathbf{h}_x \, d\mathbf{h}_y. \qquad (8)$$

To obtain tractable inference, we introduce the following factored variational posterior

$$q_{\boldsymbol{\phi}}(\mathbf{z}, \mathbf{h}_x, \mathbf{h}_y|\mathbf{x}, \mathbf{y}) =$$
$$q_{\boldsymbol{\phi}}(\mathbf{z}|\mathbf{x}; \boldsymbol{\phi}_z)q_{\boldsymbol{\phi}}(\mathbf{h}_x|\mathbf{x}; \boldsymbol{\phi}_x)q_{\boldsymbol{\phi}}(\mathbf{h}_y|\mathbf{y}; \boldsymbol{\phi}_y), \qquad (9)$$

where each factor is parameterized by a different DNN. Similarly to VCCA, we can derive a variational lower bound on the data log-likelihood for VCCA-private as (see the full derivation in Appendix B)

$$\log p_{\boldsymbol{\theta}}(\mathbf{x}, \mathbf{y}) \geq \mathcal{L}_{\text{private}}(\mathbf{x}, \mathbf{y}; \boldsymbol{\theta}, \boldsymbol{\phi}) := -D_{KL}(q_{\boldsymbol{\phi}}(\mathbf{z}|\mathbf{x})||p(\mathbf{z}))$$
$$- D_{KL}(q_{\boldsymbol{\phi}}(\mathbf{h}_x|\mathbf{x})||p(\mathbf{h}_x)) - D_{KL}(q_{\boldsymbol{\phi}}(\mathbf{h}_y|\mathbf{y})||p(\mathbf{h}_y))$$
$$+ \mathbb{E}_{q_{\boldsymbol{\phi}}(\mathbf{z}|\mathbf{x}),\, q_{\boldsymbol{\phi}}(\mathbf{h}_x|\mathbf{x})} \left[\log p_{\boldsymbol{\theta}}(\mathbf{x}|\mathbf{z}, \mathbf{h}_x)\right]$$
$$+ \mathbb{E}_{q_{\boldsymbol{\phi}}(\mathbf{z}|\mathbf{x}),\, q_{\boldsymbol{\phi}}(\mathbf{h}_y|\mathbf{y})} \left[\log p_{\boldsymbol{\theta}}(\mathbf{y}|\mathbf{z}, \mathbf{h}_y)\right]. \qquad (10)$$

VCCA-private maximizes this bound on the training set:

$$\max_{\boldsymbol{\theta}, \boldsymbol{\phi}} \frac{1}{N} \sum_{i=1}^{N} \mathcal{L}_{\text{private}}(\mathbf{x}_i, \mathbf{y}_i; \boldsymbol{\theta}, \boldsymbol{\phi}). \qquad (11)$$

As in VCCA, the last two terms of (10) can be approximated by Monte Carlo sampling. In particular, we draw samples of $\mathbf{z}$ and $\mathbf{h}_x$ from their corresponding approximate posteriors, and concatenate their samples as inputs to the DNN parameterizing $p_{\boldsymbol{\theta}}(\mathbf{x}|\mathbf{z}, \mathbf{h}_x)$. In this paper, we use simple Gaussian prior distributions for the private variables, i.e., $\mathbf{h}_x \sim \mathcal{N}(\mathbf{0}, \mathbf{I})$ and $\mathbf{h}_y \sim \mathcal{N}(\mathbf{0}, \mathbf{I})$. We leave to future work to examine the effect of more sophisticated prior distributions for the latent variables.

**Optimization**   Unlike the deep CCA objective, our objectives (4) and (11) decouple over the training samples and can be trained efficiently using stochastic gradient descent. Enabled by the reparameterization trick, unbiased gradient estimates are obtained by Monte Carlo sampling and the standard backpropagation procedure on minibatches of training samples. We apply the ADAM algorithm (Kingma & Ba, 2015) for optimizing our objectives.

### 2.2. Choice of lower bounds

In the presentation above, we have parameterized $q(\mathbf{z}|\mathbf{x})$ to obtain the VCCA and VCCA-private objectives. This is convenient when only the first view is available for downstream tasks, in which case we can directly apply $q(\mathbf{z}|\mathbf{x})$ to obtain its projection as features. One could also derive likelihood lower bounds by parameterizing the approximate posteriors $q(\mathbf{z}|\mathbf{y})$ or $q(\mathbf{z}|\mathbf{x}, \mathbf{y})$, and optimize their convex combinations for training.

Empirically, we find that using the lower bound derived from $q(\mathbf{z}|\mathbf{x})$ tends to give the best downstream task performance when only $\mathbf{x}$ is present at test time, probably because the training procedure simulates well the test scenario. Another useful objective that we will demonstrate (on the MIR-Flickr data set in Section 4.3) is the convex combination of the two lower bounds derived from $q(\mathbf{z}|\mathbf{x})$ and $q(\mathbf{z}|\mathbf{y})$ respectively:

$$\mu \tilde{\mathcal{L}}_{q(\mathbf{z}|\mathbf{x})}(\mathbf{x}, \mathbf{y}) + (1 - \mu)\tilde{\mathcal{L}}_{q(\mathbf{z}|\mathbf{y})}(\mathbf{x}, \mathbf{y}) \qquad (12)$$

where $\mu \in [0, 1]$ and $\tilde{\mathcal{L}}$ can be either the VCCA or VCCA-private objective. We refer to these variants of the objective as bi-VCCA and bi-VCCA-private. We can still work in the setting where only one of the views is available at test time. However, when both $\mathbf{x}$ and $\mathbf{y}$ are present at test time (as for the multi-modal retrieval task), we use the concatenation of projections by $q(\mathbf{z}|\mathbf{x})$ and $q(\mathbf{z}|\mathbf{y})$ as features.

## 3. Related work

Recently, there has been much interest in unsupervised deep generative models (Kingma & Welling, 2014; Rezende et al., 2014; Goodfellow et al., 2014; Gregor et al., 2015; Makhzani et al., 2016; Burda et al., 2016; Alain et al., 2016). A common motivation behind these models is that, with the expressive power of DNNs, the generative models can capture distributions for complex inputs. Additionally, if we are able to generate realistic samples from the learned distribution, we can infer that we have discovered the underlying structure of the data, which may allow us to reduce the sample complexity for learning for downstream tasks. These previous models have mostly focused on single-view data. Here we focus on the multiview setting where multiple views of the data are present for feature extraction but often only one view is available at test time (in downstream tasks).

Some recent work has explored deep generative models for (semi-)supervised learning. Kingma et al. (2014) built a generative model based on variational autoencoders (VAEs) for semi-supervised classification, where the authors model the input distribution with two set of latent variables: the class label (if it is missing) and another set that models the intra-class variabilities (styles). Sohn et al. (2015) proposed a conditional generative model for structured output prediction, where the authors explicitly model the uncertainty in the input/output using Gaussian latent variables. While there are two set of observations (input and output labels) in this work, their graphical models are different from that of VCCA.

Our work is also related to deep multi-view probabilistic models based on restricted Boltzmann machines (Srivastava & Salakhutdinov, 2014; Sohn et al., 2014). We note that these are undirected graphical models for which both inference and learning are difficult, and one typically resorts to carefully designed variational approximation and Gibbs sampling procedures for training such models. In contrast, our models only require sampling from simple, standard distributions (such as Gaussians), and all parameters can be learned end-to-end by standard stochastic gradient methods. Therefore, our models are more scalable than the previous multi-view probabilistic models.

There is also a rich literature in modeling multi-view data using the same or similar graphical models behind VCCA/VCCA-private (Shon et al., 2006; Wang, 2007; Jia et al., 2010; Salzmann et al., 2010; Virtanen et al., 2011; Memisevic et al., 2012; Damianou et al., 2012; Klami et al., 2013). Our methods differ from previous work in parameterizing the probability distributions using DNNs. This makes the model more powerful, while still having tractable objectives and efficient end-to-end training using the local reparameterization technique. We note that, unlike earlier work on probabilistic models of linear CCA (Bach & Jordan, 2005), VCCA does not optimize the same criterion, nor produce the same solution, as any linear or nonlinear CCA. However, we retain the terminology in order to clarify the connection with earlier work on probabilistic models for CCA, which we are extending with DNN models for the observations and for the variational posterior distribution approximation.

Finally, the information bottleneck (IB) method is equivalent to linear CCA for Gaussian input variables (Chechik et al., 2005). In parallel work, Alemi et al. (2017) have extended IB to DNN-parameterized densities and derived a variational lower bound of the IB objective. Interestingly, their lower bound is closely related to that of our basic VCCA, but their objective does not contain the likelihood for the first view and has a trade-off parameter for the KL divergence term.

## 4. Experimental results

In this section, we compare different multi-view representation learning methods on three tasks involving several domains: image-image, speech-articulation, and image-text. The methods we choose to compare below are closely related to ours or have been shown to have strong empirical performance under similar settings.

**CCA**: its probabilistic interpretation motivates this work.

**Deep CCA** (DCCA): see its objective in (1).

**Deep canonically correlated autoencoders** (DCCAE, Wang et al., 2015b): combines the DCCA objective and reconstruction errors of the two views.

**Multi-view autoencoder** (MVAE, Ngiam et al., 2011): see its objective in (7).

**Multi-view contrastive loss** (Hermann & Blunsom, 2014): based on the intuition that the distance between embeddings of paired examples $\mathbf{x}^+$ and $\mathbf{y}^+$ should be smaller than the distance between embeddings of $\mathbf{x}^+$ and an unmatched negative example $\mathbf{y}^-$ by a margin:

$$\min_{f,g} \frac{1}{N} \sum_i^N \max\left(0,\ m + dis\left(f(\mathbf{x}_i^+), g(\mathbf{y}_i^+)\right)\right.$$
$$\left. - dis\left(f(\mathbf{x}_i^+), g(\mathbf{y}_i^-)\right)\right),$$

where $\mathbf{y}_i^-$ is a randomly sampled view 2 example, and $m$ is a margin hyperparameter. We use the cosine distance $dis\left(\mathbf{a}, \mathbf{b}\right) = 1 - \left\langle \frac{\mathbf{a}}{\|\mathbf{a}\|}, \frac{\mathbf{b}}{\|\mathbf{b}\|} \right\rangle$.

### 4.1. Noisy MNIST dataset

We first demonstrate our algorithms on the noisy MNIST dataset used by Wang et al. (2015b). The dataset is generated using the MNIST dataset (LeCun et al., 1998), which consists of $28 \times 28$ grayscale digit images, with $60K/10K$ images for training/testing. We first linearly rescale the pixel values to the range $[0, 1]$. Then, we randomly rotate the images at angles uniformly sampled from $[-\pi/4, \pi/4]$ and the resulting images are used as view 1 inputs. For each view 1 image, we randomly select an image of the same identity (0-9) from the original dataset, add independent random noise uniformly sampled from $[0, 1]$ to each pixel, and truncate the pixel final values to $[0, 1]$ to obtain the corresponding view 2 sample. A selection of input images is given in Figure 3 (left). The original training set is further split into training/tuning sets of size $50K/10K$. The data generation process ensures that the digit identity is the only common variable underlying both views.

To evaluate the amount of class information extracted by different methods, after unsupervised learning of latent representations, we reveal the labels and train a linear SVM on the projected view 1 training data (using the one-versus-all scheme), and use it to classify the projected test set. This experiment simulates the typical usage of multi-view learning/testing, which is to extract useful representations for downstream discriminative tasks.

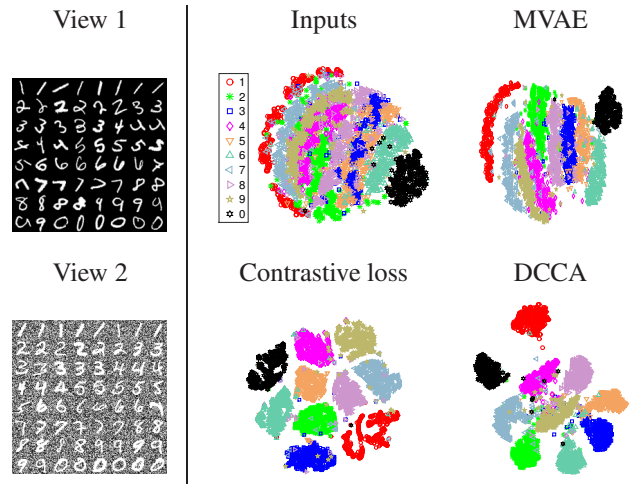

*Figure 3.* Left: Selection of view 1 images (top) and their corresponding view 2 images (bottom) from noisy MNIST. Right: 2D t-SNE visualization of features learned by previous methods.

Note that this synthetic dataset perfectly satisfies the multi-view assumption that the two views are independent given the class label, so the latent representation should contain precisely the class information. This is indeed achieved by CCA-based and contrastive loss-based multi-view approaches. In Figure 3 (right), we show 2D t-SNE (van der Maaten & Hinton, 2008) visualizations of the original view 1 inputs and view 1 projections by various deep multi-view methods.

We use DNNs with 3 hidden layers of 1024 rectified linear units (ReLUs, Nair & Hinton, 2010) each to parameterize the VCCA/VCCA-private distributions $q_\phi(\mathbf{z}|\mathbf{x})$, $p_\theta(\mathbf{x}|\mathbf{z})$, $p_\theta(\mathbf{y}|\mathbf{z})$, $q_\phi(\mathbf{h}_x|\mathbf{x})$, $q_\phi(\mathbf{h}_y|\mathbf{y})$. The capacities of these networks are the same as those of their counterparts in DCCA and DCCAE from Wang et al. (2015b). The reconstruction networks $p_\theta(\mathbf{x}|\mathbf{z})$ or $p_\theta(\mathbf{x}|\mathbf{z}, \mathbf{h}_x)$ model each pixel of $\mathbf{x}$ as an independent Bernoulli variable and parameterize its mean (using a sigmoid activation); $p_\theta(\mathbf{y}|\mathbf{z})$ and $p_\theta(\mathbf{y}|\mathbf{z}, \mathbf{h}_y)$ model $\mathbf{y}$ with diagonal Gaussians and parameterize the mean (using a sigmoid activation) and standard deviation for each pixel dimension. We tune the dimensionality $d_z$ over $\{10, 20, 30, 40, 50\}$, and fix $d_{h_x} = d_{h_y} = 30$ for VCCA-private. We select the hyperparameter combination that yields the best SVM classification accuracy on the projected tuning set, and report the corresponding accuracy on the projected test set.

**Learning compact representations** We add dropout (Srivastava et al., 2014) to all intermediate layers and the input layers and find it to be very useful, with most of the gain coming from dropout applied to the samples of **z**, **h**$_x$ and **h**$_y$. Dropout encourages each latent dimension to reconstruct the inputs well in the absence of other dimensions, and therefore avoids learning co-adapted features; dropout has also been found to be useful in other deep generative models (Sohn et al., 2015). Intuitively, in VCCA-private dropout also helps to prevent the degenerate situation where the pathways **x** → **h**$_x$ → **x** and **y** → **h**$_y$ → **y** achieve good reconstruction while ignoring **z** (e.g., by setting it to a constant). We have experimented with the orthogonal penalty of Bousmalis et al. (2016) which minimizes the correlation between shared and private variables (see their eqn. 5), but it is outperformed by dropout in our experiments. In fact, with dropout, the correlation between the two blocks of variables decreases without using the orthogonal penalty; see Appendix C for experimental results on this phenomenon. We use the same dropout rate for all layers and tune it over $\{0, 0.1, 0.2, 0.4\}$.

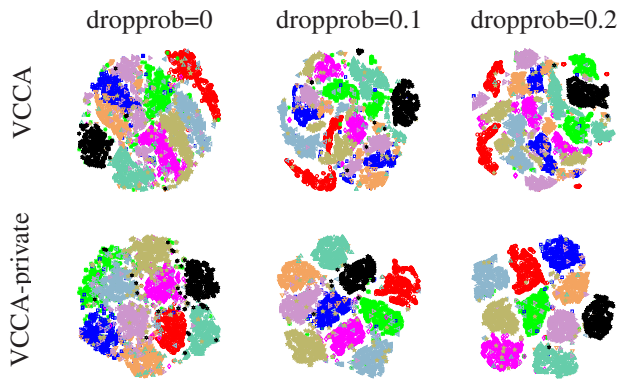

Figure 4. 2D t-SNE visualizations of the extracted shared variables **z** on noisy MNIST test data by VCCA (top row) and VCCA-private (bottom row) for different dropout rates. Here $d_z = 40$.

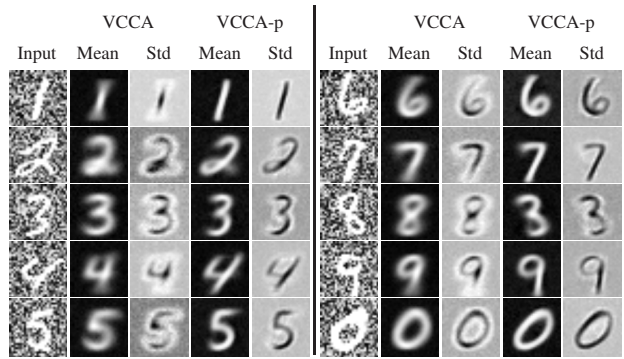

Figure 5. Sample reconstruction of view 2 images from the noisy MNIST test set by VCCA and VCCA-private.

Figure 4 shows 2D t-SNE visualizations of the common variables **z** learned by VCCA and VCCA-private. In general, VCCA/VCCA-private separate the classes well; dropout significantly improves the performance of both VCCA and VCCA-private, with the latter slightly outperforming the former. While such class separation can also be achieved by DCCA/contrastive loss, these methods can not naturally generate samples in the input space. Recall that such separation is not achieved by MVAE (Figure 3).

**The effect of private variables on reconstructions** Figure 5 (columns 2 and 3 in each panel) shows sample reconstructions (mean and standard deviation) by VCCA for the view 2 images from the test set; more examples are provided in Appendix D. We observe that for each input, the mean reconstruction of **y**$_i$ by VCCA is a prototypical image of the same digit, regardless of the individual style in **y**$_i$. This is to be expected, as **y**$_i$ contains an arbitrary im-age of the same digit as **x**$_i$, and the variation in background noise in **y**$_i$ does not appear in **x**$_i$ and can not be reflected in $q_\phi(\mathbf{z}|\mathbf{x})$; thus the best way for $\mathbf{p}_\theta(\mathbf{y}|\mathbf{z})$ to model **y**$_i$ is to output a prototypical image of that class to achieve on average small reconstruction error. On the other hand, since **y**$_i$ contains little rotation of the digits, this variation is suppressed to a large extent in $q_\phi(\mathbf{z}|\mathbf{x})$.

Figure 5 (columns 4 and 5 in each panel) shows sample reconstructions by VCCA-private for the same set of view 2 images. With the help of private variables **h**$_y$ (as part of the input to $p_\theta(\mathbf{y}|\mathbf{z}, \mathbf{h}_y)$), the model does a much better job in reconstructing the styles of **y**. And by disentangling the private variables from the shared variables, $q_\phi(\mathbf{z}|\mathbf{x})$ achieves even better class separation than VCCA does. We also note that the standard deviation of the reconstruction is low within the digit and high outside the digit, implying that $p_\theta(\mathbf{y}|\mathbf{z}, \mathbf{h}_y)$ is able to separate the background noise from the digit image.

**Disentanglement of private/shared variables** In Figure 6 we provide 2D t-SNE embeddings of the shared variables **z** (top row) and private variables **h**$_x$ (bottom row) learned by VCCA-private. In the embedding of **h**$_x$, digits with different identities but the same rotation are mapped close together, and the rotation varies smoothly from left to right, confirming that the private variables contain little class information but mainly style information.

Finally, in Table 1 we give the test error rates of linear SVMs applied to the features learned with different models. VCCA-private is comparable in performance to the best previous approach (DCCAE), while having the advantage that it can also generate. See Appendix E for samples of generated images using VCCA-private.

### 4.2. XRMB speech-articulation dataset

We now consider the task of learning acoustic features for speech recognition. We use data from the Wisconsin X-ray microbeam (XRMB) corpus (Westbury, 1994), which con-tains simultaneously recorded speech and articulatory measurements from 47 American English speakers. We follow the setup of Wang et al. (2015a;b) and use the learned features for speaker-independent phonetic recognition.[2] The two input views are standard 39D acoustic features (13 mel frequency cepstral coefficients (MFCCs) and their first and second derivatives) and 16D articulatory features (horizontal/vertical displacement of 8 pellets attached to several parts of the vocal tract), each then concatenated over a 7-frame window around each frame to incorporate context. The speakers are split into disjoint sets of 35/8/2/2 speakers for feature learning/recognizer training/tuning/testing. The 35 speakers for feature learning are fixed; the remaining 12 are used in a 6-fold experiment (recognizer training on 8 speakers, tuning on 2 speakers, and testing on the remaining 2 speakers). Each speaker has roughly $50K$ frames.

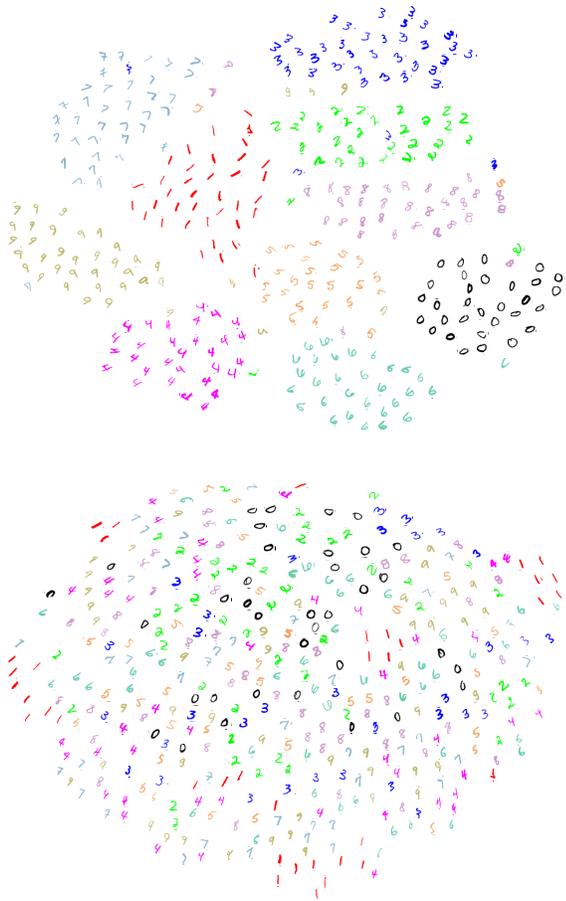

*Figure 6.* 2D *t*-SNE embedding of the shared variables $\mathbf{z} \in \mathbb{R}^{40}$ (top) and private variables $\mathbf{h}_x \in \mathbb{R}^{30}$ (bottom).

*Table 1.* Performance of different features for downstream tasks: Classification error rates of linear SVMs on noisy MNIST, mean phone error rate (PER) over 6 folds on XRMB, and mean average precision (mAP) for unimodal retrieval on Flickr. * Results from Wang et al. (2015b). [+] Results from Wang & Livescu (2016).

| Method | MNIST Error (%) | XRMB PER (%, ↓) | Flickr mAP (↑) |
|---|---|---|---|
| Original inputs | 13.1* | 37.6[+] | 0.480 |
| CCA | 19.1* | 29.4[+] | 0.529 |
| DCCA | 2.9* | 25.4[+] | 0.573 |
| DCCAE | 2.2* | 25.4 | 0.573 |
| Contrastive | 2.7 | 24.6 | 0.565 |
| MVAE (orig) | 11.7* | 29.4 | 0.477 |
| MVAE-var | - | - | 0.595 |
| VCCA | 3.0 | 28.0 | 0.605 |
| VCCA-private | 2.4 | 25.2 | 0.615 |
| bi-VCCA | - | - | 0.606 |
| bi-VCCA-private | - | - | 0.626 |

We remove the per-speaker mean and variance of the articulatory measurements for each training speaker, and remove the mean of the acoustic measurements for each utterance. All learned feature types are used in a "tandem" speech recognizer (Hermansky et al., 2000), i.e., they are appended to the original 39D features and used in a standard hidden Markov model (HMM)-based recognizer with Gaussian mixture observation distributions.

Each algorithm uses up to 3 ReLU hidden layers, each of 1500 units, for the projection and reconstruction mappings. For VCCA/VCCA-private, we use Gaussian observation models as the inputs are real-valued. In contrast to the MNIST experiments, we do not learn the standard deviations of each output dimension on training data, as this leads to poor downstream task performance. Instead, we use isotropic covariances for each view, and tune the standard deviations by grid search. The best model uses a smaller standard deviation (0.1) for view 2 than for view 1 (1.0), effectively putting more emphasis on the reconstruction of articulatory measurements. Our best-performing VCCA model uses $d_z = 70$, while the best-performing VCCA-private model uses $d_z = 70$ and $d_{h_x} = d_{h_y} = 10$.

The mean phone error rates (PER) over 6 folds obtained by different algorithms are given in Table 1. Our methods achieve competitive performance in comparison to previous deep multi-view methods.

### 4.3. MIR-Flickr dataset

Finally, we consider the task of learning cross-modality features for topic classification on the MIR-Flickr database (Huiskes & Lew, 2008). The Flickr database con-tains 1 million images accompanied by user tags, among which 25000 images are labeled with 38 topic classes (each image may be categorized as multiple topics). We use the same image and text features as in previous work (Srivastava & Salakhutdinov, 2014; Sohn et al., 2014): the image feature vector is a 3857-dimensional real-valued vector of handcrafted features, while the text feature vector is a 2000-dimensional binary vector of frequent tags.

---

[2]As in (Wang & Livescu, 2016), we use the Kaldi toolkit (Povey et al., 2011) for feature extraction and recognition with hidden Markov models. Our results do not match Wang et al. (2015a;b) (who instead used the HTK toolkit (Young et al., 1999)) for the same types of features, but the relative results are consistent.

Following the same protocol as Sohn et al. (2014), we train multi-view representations using the unlabelled data,[3] and use projected image features of the labeled data (further divided into splits of 10000/5000/10000 samples for training/tuning/testing) for training and evaluating a classifier that predicts the topic labels, corresponding to the unimodal query task in Srivastava & Salakhutdinov (2014); Sohn et al. (2014). For each algorithm, we select the model achieving the highest mean average precision (mAP) on the validation set, and report its performance on the test set.

Each algorithm uses up to 4 ReLU hidden layers, each of 1024 units, for the projection and reconstruction mappings. For VCCA/VCCA-private, we use Gaussian observation models with isotropic covariance for image features, with standard deviation tuned by grid search, and a Bernoulli model for text features. For comparison with multi-view autoencoders (MVAE), we considered both the original MVAE objective (7) with $\ell_2$ reconstruction errors and a new variant (MVAE-var below) with a cross-entropy reconstruction loss on the text view; MVAE-var matches the reconstruction part of the VCCA objective when using the Bernoulli model for the text view. In this experiment, we found it helpful to tune an additional trade-off parameter for the text-view likelihood (cross-entropy); the best VCCA/VCCA-private models prefer a large trade-off parameter ($10^4$), emphasizing the reconstruction of the sparse text-view inputs. Our best-performing VCCA model uses $d_z = 1024$, while the best performing VCCA-private model uses $d_z = 1024$ and $d_{h_x} = d_{h_y} = 16$.

Furthermore, we have explored the bi-VCCA/bi-VCCA-private objectives (12) with intermediate values of $\mu$ (recall that $\mu = 1$ gives the usual lower bound derived from $q(\mathbf{z}|\mathbf{x})$), and found that the best unimodal retrieval performance is achieved at $\mu = 0.8$ and $\mu = 0.5$ for bi-VCCA and bi-VCCA-private respectively (although $\mu = 1$ already works well). This shows that the second lower bound can be useful in regularizing the reconstruction networks (which are shared by the two lower bounds). We present empirical analysis of $\mu$ in Appendix F.

As shown in Table 1, VCCA and VCCA-private achieve higher mAPs than other methods considered here, as well as the previous state-of-the-art mAP result of 0.607 achieved by the multi-view RBMs (MVRBM) of Sohn et al. (2014) under the same setting. Unlike in the MNIST and XRMB tasks, we observe sizable gains over DCCAE and contrastive losses. We conjecture that this is expected in tasks, like MIR-Flickr, where one of the views is sparse (in the case of MIR-Flickr, because there are many more potential textual tags than are actually used), so contrastive losses may have trouble finding appropriate negative examples. VCCA and its variants are also much easier to train than prior state-of-the-art methods. In addition, if both views are present at test time, we can use concatenated projections $q(\mathbf{z}|\mathbf{x})$ and $q(\mathbf{z}|\mathbf{y})$ from bi-VCCA-private (12) and perform multimodal retrieval; taking this approach with $\mu = 0.5$, we achieve a mAP of 0.687, comparable to that of Sohn et al. (2014).

## 5. Conclusions

We have proposed variational canonical correlation analysis (VCCA), a deep generative method for multi-view representation learning. Our method embodies a natural idea for multi-view learning: the multiple views can be generated from a small set of shared latent variables. VCCA is parameterized by DNNs and can be trained efficiently by backpropagation, and is therefore scalable. We have also shown that, by modeling the private variables that are specific to each view, the VCCA-private variant can disentangle shared/private variables and provide higher-quality features and reconstructions. When using the learned representations in downstream prediction tasks, VCCA and its variants are competitive with or improve upon prior state-of-the-art results, while being much easier to train.[4]

Future work includes exploration of additional prior distributions such as mixtures of Gaussians or discrete random variables, which may enforce clustering in the latent space and in turn work better for discriminative tasks. In addition, we have thus far used a standard black-box variational inference technique with good scalability; recent developments in variational inference (Rezende & Mohamed, 2015; Tran et al., 2016) may improve the expressiveness of the model and the features. We will also explore other observation models, including replacing the auto-encoder objective with that of adversarial networks (Goodfellow et al., 2014; Makhzani et al., 2016; Chen et al., 2016).

---

[3] As in Sohn et al. (2014), we exclude about 250000 samples that contain fewer than two tags.

[4] Our implementation is available at www.

## A. Derivation of the variational lower bound for VCCA

We can derive a lower bound on the marginal data likelihood using $q_\phi(\mathbf{z}|\mathbf{x})$:

$$\log p_{\boldsymbol{\theta}}(\mathbf{x}, \mathbf{y})$$

$$= \log p_{\boldsymbol{\theta}}(\mathbf{x}, \mathbf{y}) \int q_\phi(\mathbf{z}|\mathbf{x}) d\mathbf{z} = \int \log p_{\boldsymbol{\theta}}(\mathbf{x}, \mathbf{y}) q_\phi(\mathbf{z}|\mathbf{x}) d\mathbf{z}$$

$$= \int q_\phi(\mathbf{z}|\mathbf{x}) \left( \log \frac{q_\phi(\mathbf{z}|\mathbf{x})}{p_{\boldsymbol{\theta}}(\mathbf{z}|\mathbf{x}, \mathbf{y})} + \log \frac{p_{\boldsymbol{\theta}}(\mathbf{x}, \mathbf{y}, \mathbf{z})}{q_\phi(\mathbf{z}|\mathbf{x})} \right) d\mathbf{z}$$

$$= D_{KL}(q_\phi(\mathbf{z}|\mathbf{x})||p_{\boldsymbol{\theta}}(\mathbf{z}|\mathbf{x}, \mathbf{y})) + \mathbb{E}_{q_\phi(\mathbf{z}|\mathbf{x})}\left[ \log \frac{p_{\boldsymbol{\theta}}(\mathbf{x}, \mathbf{y}, \mathbf{z})}{q_\phi(\mathbf{z}|\mathbf{x})} \right]$$

$$\geq \mathbb{E}_{q_\phi(\mathbf{z}|\mathbf{x})}\left[ \log \frac{p_{\boldsymbol{\theta}}(\mathbf{x}, \mathbf{y}, \mathbf{z})}{q_\phi(\mathbf{z}|\mathbf{x})} \right]$$

$$= \mathcal{L}(\mathbf{x}, \mathbf{y}; \boldsymbol{\theta}, \boldsymbol{\phi}) \tag{13}$$

where we used the fact that KL divergence is nonnegative in the last step. As a result, $\mathcal{L}(\mathbf{x}, \mathbf{y}; \boldsymbol{\theta}, \boldsymbol{\phi})$ is a lower bound on the data log-likelihood $\log_{\boldsymbol{\theta}} p(\mathbf{x}, \mathbf{y})$.

Substituting (2) into (13), we have

$$\mathcal{L}(\mathbf{x}, \mathbf{y}; \boldsymbol{\theta}, \boldsymbol{\phi})$$

$$= \int q_\phi(\mathbf{z}|\mathbf{x}) \left[ \log \frac{p(\mathbf{z})}{q_\phi(\mathbf{z}|\mathbf{x})} + \log p_{\boldsymbol{\theta}}(\mathbf{x}|\mathbf{z}) + \log p_{\boldsymbol{\theta}}(\mathbf{y}|\mathbf{z}) \right] d\mathbf{z}$$

$$= - D_{KL}(q_\phi(\mathbf{z}|\mathbf{x})||p(\mathbf{z})) $$
$$\quad\quad + \mathbb{E}_{q_\phi(\mathbf{z}|\mathbf{x})}\left[ \log p_{\boldsymbol{\theta}}(\mathbf{x}|\mathbf{z}) + \log p_{\boldsymbol{\theta}}(\mathbf{y}|\mathbf{z}) \right]$$

as desired.

## B. Derivation of the variational lower bound for VCCA-private

Similar to the derivation for VCCA, we have

$$\log p_{\boldsymbol{\theta}}(\mathbf{x}, \mathbf{y})$$

$$= \log \iiint p_{\boldsymbol{\theta}}(\mathbf{x}, \mathbf{y}, \mathbf{z}, \mathbf{h}_x, \mathbf{h}_y) d\mathbf{z} \, d\mathbf{h}_x \, d\mathbf{h}_y$$

$$\geq \iiint q_\phi(\mathbf{z}, \mathbf{h}_x, \mathbf{h}_y|\mathbf{x}, \mathbf{y}) \log \frac{p_{\boldsymbol{\theta}}(\mathbf{x}, \mathbf{y}, \mathbf{z}, \mathbf{h}_x, \mathbf{h}_y)}{q_\phi(\mathbf{z}, \mathbf{h}_x, \mathbf{h}_y|\mathbf{x}, \mathbf{y})} d\mathbf{z} \, d\mathbf{h}_x \, d\mathbf{h}_y$$

$$= \iiint q_\phi(\mathbf{z}, \mathbf{h}_x, \mathbf{h}_y|\mathbf{x}, \mathbf{y}) \left[ \log \frac{p(\mathbf{z})}{q_\phi(\mathbf{z}|\mathbf{x})} + \log \frac{p(\mathbf{h}_x)}{q_\phi(\mathbf{h}_x|\mathbf{x})} \right.$$

$$\quad\quad + \log \frac{p(\mathbf{h}_y)}{q_\phi(\mathbf{h}_y|\mathbf{y})} + \log p_{\boldsymbol{\theta}}(\mathbf{x}|\mathbf{z}, \mathbf{h}_x)$$

$$\quad\quad \left. + \log p_{\boldsymbol{\theta}}(\mathbf{y}|\mathbf{z}, \mathbf{h}_y) \right] d\mathbf{z} \, d\mathbf{h}_x \, d\mathbf{h}_y$$

$$= - D_{KL}(q_\phi(\mathbf{z}|\mathbf{x})||p(\mathbf{z})) - D_{KL}(q_\phi(\mathbf{h}_x|\mathbf{x})||p(\mathbf{h}_x))$$

$$\quad\quad - D_{KL}(q_\phi(\mathbf{h}_y|\mathbf{y})||p(\mathbf{h}_y))$$

$$\quad\quad + \iint q_\phi(\mathbf{z}|\mathbf{x}) q_\phi(\mathbf{h}_x|\mathbf{x}) \log p_{\boldsymbol{\theta}}(\mathbf{x}|\mathbf{z}, \mathbf{h}_x) d\mathbf{z} \, d\mathbf{h}_x$$

$$\quad\quad + \iint q_\phi(\mathbf{z}|\mathbf{x}) q_\phi(\mathbf{h}_y|\mathbf{y}) \log p_{\boldsymbol{\theta}}(\mathbf{y}|\mathbf{z}, \mathbf{h}_y) d\mathbf{z} \, d\mathbf{h}_y$$

$$= \mathcal{L}_{\text{private}}(\mathbf{x}, \mathbf{y}; \boldsymbol{\theta}, \boldsymbol{\phi}). \tag{14}$$

## C. Analysis of orthogonality between shared and private variables

As mentioned in the main text, we would like to learn disentangled representations for the shared and private variables. Thus ideally, the shared and private variables should be as orthogonal to each other as possible. Let $\mathbf{Z}$ and $\mathbf{H}_x$ be matrics whose rows contain the means of $q_\phi(\mathbf{z}|\mathbf{x})$ and $q_\phi(\mathbf{h}_x|\mathbf{x})$ respectively for a set of samples. We use the following score to quantitatively measure the orthogonality between the shared and private variables:

$$\lambda_{\mathbf{Z}\perp\mathbf{H}_x} = \frac{\|\mathbf{H}_x^\top \mathbf{Z}\|_F^2}{\|\mathbf{H}_x\|_F^2 \cdot \|\mathbf{Z}\|_F^2} \qquad (15)$$

where $\|\cdot\|_F$ is the Frobenius norm. The score is zero when the two variables are orthogonal to each other. On the other hand, when the two variables are almost identical (with the same dimensionality), the score has value $1$.

On the noisy MNIST dataset, we evaluate the orthogonality score between shared and private variables (view 1 and view 2, respectively) at every epoch for the entire validation set, and compare orthogonality scores from models trained with and without dropout. As shown in Figure 7, the model trained with dropout achieves better orthogonality between shared and private variables from both views. In contrast, the orthogonality scores are clearly higher for model trained without dropout. In this case, it is quite likely that the model (with millions of parameters) overfits to the data (noisy MNIST has only 50,000 training samples) by ignoring the shared variables.

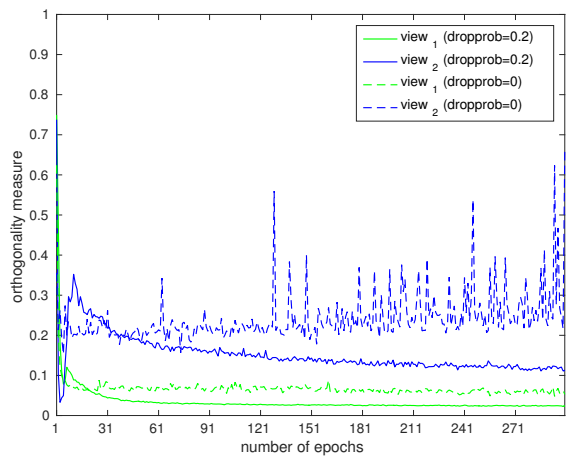

*Figure 7.* Orthogonality score curves on the noisy MNIST validation set.

## D. Additional reconstruction results of noisy MNIST

In Figure 8 we provide additional examples to demonstrate the effect of private variables in reconstruction.

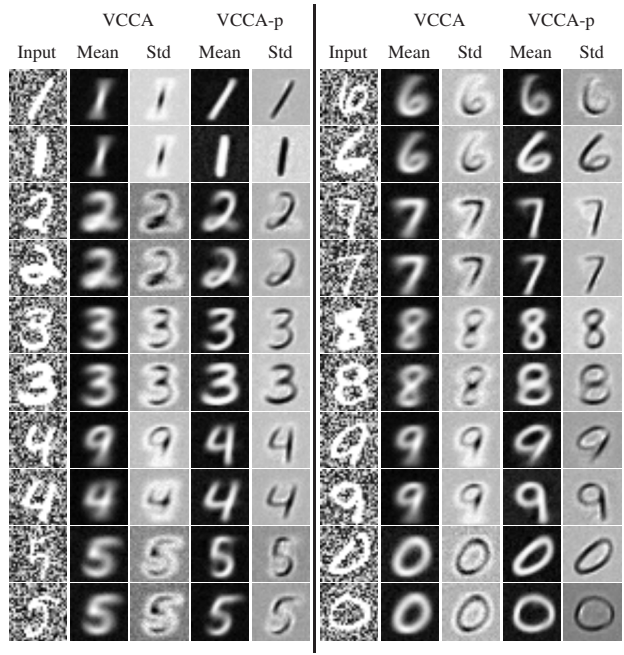

*Figure 8.* Sample reconstruction of view 2 images from the noisy MNIST test set by VCCA and VCCA-private.

## E. Additional generation examples for noisy MNIST

To better demonstrate the role of private variables, we perform manifold traversal along the private dimensions while fixing the shared dimensions on noisy MNIST. Specifically, given an input MNIST digit $\mathbf{x}$, we first infer the shared variables $\mathbf{z} \sim q_\phi(\mathbf{z}|\mathbf{x})$. Rather than reconstructing the input as has been done in the reconstruction experiment in the main text, we attempt to augment the input by generating samples $\mathbf{x}' \sim p_\theta(\mathbf{x}|\mathbf{z}, \mathbf{h}_x)$ with diverse $\mathbf{h}_x \sim p(\mathbf{h}_x)$.

As we can see in Figure 9, with fixed shared variables, the generated samples almost always have the same identity (class label) as the input digit. However, the generated samples are quite diverse in terms of orientation, which is the main source of variation in the first view.

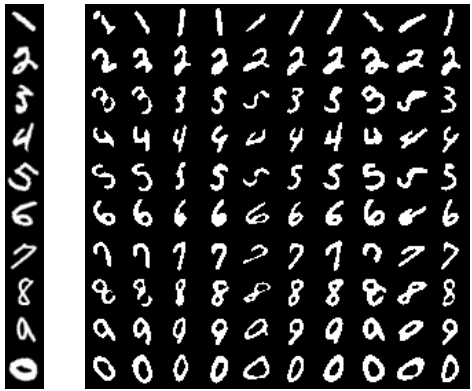

Figure 9. Generated samples from VCCA-private with diverse private variables. Input images are shown in the first column; generated samples are shown in the 10-by-10 matrix on the right. All the digits (including samples and input) in the same row share a common $\mathbf{z}$, while all the digits in the same column share a common $\mathbf{h}_x$.

## F. Empirical analysis of bi-VCCA and bi-VCCA-private on MIR-Flickr

In Table 2, we present the mAP performance of the bi-VCCA and bi-VCCA-private objectives for different values of $\mu$, on the MIR-Flickr validation set for unimodal retrieval. Recall that these objectives reduce to VCCA and VCCA-private for $\mu = 1$.

As we can see in Table 2, the improvement produced by different $\mu$ is non-trivially important for models with private variables. This is quite interesting since it indicates that optimizing the lower bound derived from $q_\phi(\mathbf{z}|\mathbf{y})$ can lead to a better $q_\phi(\mathbf{z}|\mathbf{x})$. Intuitively, when the observations from one view are ambiguous enough, observations from the other views may be more helpful. However, we do observe the same behavior in the MNIST experiment, since inferring the identity (class label) when the digits are rotated or corrupted to some degree is still possible.

Table 2. Mean average precision (mAP) of the bi-VCCA and bi-VCCA-private features on MIR-Flickr validation set for different values of $\mu$.

| Objective | $\mu = 1$ | $\mu = 0.8$ | $\mu = 0.5$ | $\mu = 0.2$ |
|---|---|---|---|---|
| bi-VCCA | 0.597 | 0.601 | 0.599 | 0.599 |
| bi-VCCA-private | 0.609 | 0.617 | 0.617 | 0.610 |